%% file: main.tex
\documentclass[10pt,twocolumn,letterpaper]{article}

\usepackage{cvpr}              

\input{preamble}

\definecolor{cvprblue}{rgb}{0.21,0.49,0.74}
\usepackage[pagebackref,breaklinks,colorlinks,allcolors=cvprblue]{hyperref}

\def\paperID{14459} 
\def\confName{CVPR}
\def\confYear{2026}

\title{PhyCo: Learning Controllable Physical Priors for Generative Motion}

\author{
Sriram Narayanan\textsuperscript{1,2} \quad
Ziyu Jiang\textsuperscript{2}\quad
Srinivasa G. Narasimhan\textsuperscript{1}\quad
Manmohan Chandraker\textsuperscript{2,3}\\[3mm]
\textnormal{
\textsuperscript{1}Carnegie Mellon University\quad
\textsuperscript{2}NEC Labs America \quad
\textsuperscript{3}UC San Diego
}\\[1mm]
\href{https://phyco-video.github.io/}{\texttt{phyco-video.github.io}}
}

\begin{document}
\input{figures/teaser}

\input{sec/0_abstract}    
\input{sec/1_intro}

\input{sec/2_related_work}
\input{sec/3_1_simdata}

\input{sec/3_2_method}

\input{sec/4_results}
\input{sec/5_conclusion}
\paragraph{\textbf{Acknowledgements:}} This work was partially conducted during Sriram’s internship at NEC Labs America, and was supported in part by NSF grants IIS-2107236 and IIS-2513219. We also thank Kausik Sivakumar and Yug Ajmera for their insightful discussions.

{
    \small
    \bibliographystyle{ieeenat_fullname}
    \bibliography{main}
}

\input{sec/X_suppl}

\end{document}

%% file: preamble.tex
\usepackage{booktabs}
\usepackage{tabularx}
\usepackage{array}
\usepackage{listings}
\usepackage{microtype}

\usepackage{amssymb}
\usepackage{makecell}
\usepackage{multirow}
\usepackage[table]{xcolor}
\usepackage{pifont} 
\newcommand{\cmark}{\ding{51}} 
\newcommand{\xmark}{\ding{55}} 

\definecolor{rowhl}{RGB}{238,245,255} 

\usepackage[most]{tcolorbox}
\usepackage{xcolor}

\usepackage[table]{xcolor}
\definecolor{best}{RGB}{255,220,220}    
\definecolor{second}{RGB}{255,245,204}  

\usepackage[normalem]{ulem} 

\newtcolorbox{vlmprompts}[2][]{
  breakable,
  colback=gray!2,
  colframe=black,
  title={#2},
  fonttitle=\bfseries,
  boxsep=1ex,
  label={#1},
}

%% file: figures/teaser.tex
\twocolumn[{%
\renewcommand\twocolumn[1][]{#1}%
\maketitle
\begin{center}
    \centering
    \captionsetup{type=figure}
    \includegraphics[trim={0 58cm 0cm 0}, clip, width=1.0\linewidth]{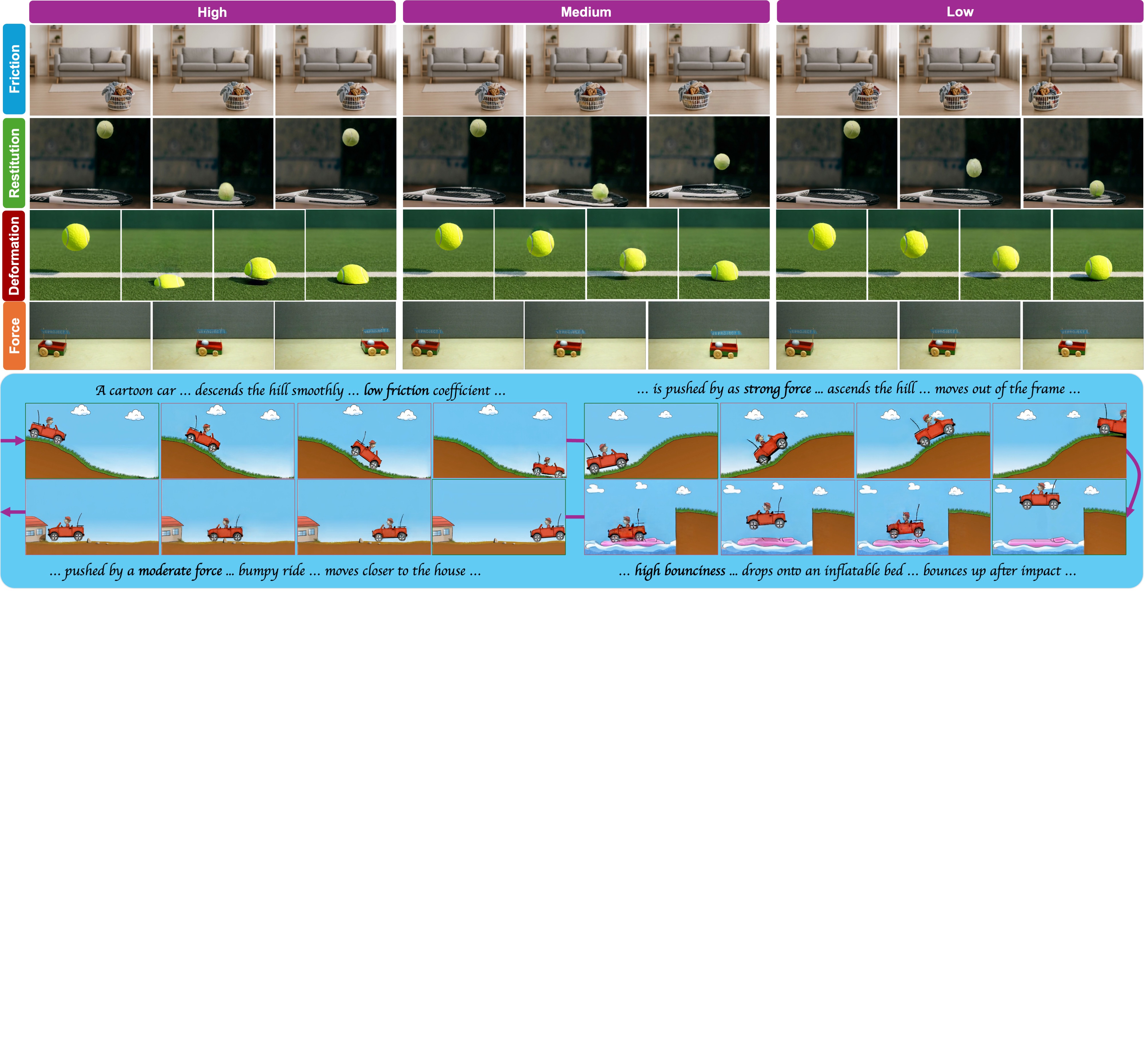}
    \captionof{figure}{
    PhyCo generates videos conditioned on spatial physical property maps, producing motion consistent with real-world dynamics. Top: Continuous control over individual attributes of friction, restitution, deformation, and applied force yields smooth and physically meaningful variations. Bottom: Strong compositional generalization enables coherent motion in stylized scenes by combining multiple attributes (e.g., force + friction, restitution + deformation), even beyond the simulation domain.
    }
    \label{fig:teaser}
\end{center}
}]

%% file: sec/0_abstract.tex
\begin{abstract}
Modern video diffusion models excel at appearance synthesis but still struggle with physical consistency: objects drift, collisions lack realistic rebound, and material responses seldom match their underlying properties. We present PhyCo, a framework that introduces continuous, interpretable, and physically grounded control into video generation. Our approach integrates three key components: (i) a large-scale dataset of over 100K photorealistic simulation videos where friction, restitution, deformation, and force are systematically varied across diverse scenarios; (ii) physics-supervised fine-tuning of a pretrained diffusion model using a ControlNet conditioned on pixel-aligned physical property maps; and (iii) VLM-guided reward optimization, where a fine-tuned vision–language model evaluates generated videos with targeted physics queries and provides differentiable feedback. This combination enables a generative model to produce physically consistent and controllable outputs through variations in physical attributes—without any simulator or geometry reconstruction at inference. On the Physics-IQ benchmark, PhyCo significantly improves physical realism over strong baselines, and human studies confirm clearer and more faithful control over physical attributes. Our results demonstrate a scalable path toward physically consistent, controllable generative video models that generalize beyond synthetic training environments. 
\end{abstract}

%% file: sec/1_intro.tex
\vspace{-0.8cm}
\section{Introduction}

Understanding and generating physically grounded behaviors is a core challenge in building intelligent visual models. Humans effortlessly infer how objects react to forces and slide, bounce, or deform when interacting with other surfaces, but it remains challenging for video generation models. While modern diffusion-based video generators excel at synthesizing realistic textures, lighting and motion continuity, they often violate the basic laws of physics: objects hover or fall too slowly under gravity, collisions occur without rebound and soft objects fail to deform realistically~\cite{physicsiq}. Importantly, despite the large variety of data on which foundational video diffusion models are trained, it remains difficult to controllably generate variations in physical properties.

This work takes a step towards bridging the above gaps between visual and physical realism and control. Impressive advances have been made to address this gap in recent works that integrate physical simulation with generative models \cite{physgen,li2025wonderplay,physdreamer}. But they depend on explicit solvers, such as rigid-body dynamics in PhysGen \cite{physgen} and MPM-based optimization in PhysDreamer \cite{physdreamer} and hybrid simulation in WonderPlay \cite{li2025wonderplay}. While this leads to fine-grained motion coherence, the need for reconstructed 3D geometry or predefined materials during inference limits scalability and generalization. 
Similarly effective advances have been achieved through implicit approaches to guide motion without explicit simulation, such as PhysCtrl \cite{physctrl2025}, VLIPP \cite{vlipp}, PISA \cite{li2025pisaexperimentsexploringphysics} and ForcePrompting \cite{forceprompting}, that embed physical cues through learned or language-driven priors using trajectory generation, vision-language reasoning, gravity-based supervision, or force-conditioned prompting. But while improving semantic consistency, they lack continuous control over a diversity of underlying physical properties.


In contrast, we present PhyCo, a framework that endows generative video models with continuous and interpretable physical property conditioning (Fig.~\ref{fig:teaser}). Instead of merely conditioning on external guidance, we explicitly train video diffusion models to represent and manipulate physical properties such as friction, restitution, deformation, and applied force. This enables controllable synthesis of physically consistent motion and interactions across diverse materials and contact conditions purely through generative modeling, without requiring geometric reconstruction or simulator feedback at inference. At the same time, it offers quantitative control over motion behavior and aligns naturally with representations used in physics simulators, allowing direct supervision and interpretable manipulation. 

We achieve these distinctions through three novel contributions. First, we introduce a large-scale, multi-scenario dataset of 100K physically grounded simulation videos with continuous physical property annotations that disentangle visual appearance from underlying physics. The dataset -- built on Kubric \cite{greff2021kubric} with PyBullet \cite{coumans2016pybullet} for physics and Blender \cite{blender} for rendering -- spans diverse materials, interactions and views, covering multiple physical regimes from rigid collisions to deformable impacts, to provide a structured foundation for learning physically meaningful dynamics that generalize beyond simulation. 
Second, we propose physics-supervised fine-tuning of a pretrained diffusion backbone (Cosmos-Predict2 \cite{nvidia2025cosmosworldfoundationmodel}) using a ControlNet \cite{controlnet} architecture that injects spatially aligned physical property maps. Third, we introduce VLM-guided reward optimization, where a fine-tuned vision-language model (VLM) evaluates generated videos through targeted physics questions, providing differentiable rewards that encourage physically plausible behavior. This combination of explicit conditioning and semantic feedback enables controllable, interpretable, and physically consistent video generation that generalizes from simulation-rich training to real-world scenarios without any simulator or handcrafted physical modeling at inference time.

In extensive evaluations on the Physics-IQ benchmark~\cite{physicsiq} and human preference studies, our approach consistently outperforms prior video models in both  physical realism and controllable variation. Moreover, it generalizes to unseen materials, forces and interactions, with compositionality across variations, demonstrating that embedding physical property priors offers a scalable path toward controllable, physically consistent \cite{liu2024frechetvideomotiondistance} generative world models.

%% file: sec/2_related_work.tex
\input{figures/pipeline}
\input{tables/dataset_compare2}

\section{Related Work}


\noindent \textbf{Physics Rich Datasets.}
A variety of benchmarks and datasets have been proposed to study object dynamics and to evaluate the predictive capabilities of deep models. However, most existing datasets are constrained in terms of physical property diversity, scene realism, and coverage of complex interactions. As a result, they often fall out of distribution for today's powerful generative video models.
They also fall short of the requirements posed by modern generative video models, which demand richer, more diverse, and physically grounded supervision. Table~\ref{tab:dataset_comparison} summarizes several representative physics-focused datasets. Although the Force-Prompting dataset~\cite{forceprompting} achieves high photorealism, it remains limited in both scene diversity and annotated physical properties.
These limitations underscore the need for large-scale, physically grounded datasets that can better align generative video models with the principles of intuitive physics and enhance their ability to produce physically plausible dynamics.

\vspace{0.2cm}
\noindent \textbf{Controllable Video Generation.}
Recent advances in video diffusion models have sparked growing interest in achieving fine-grained control and physical grounding in video generation. A large body of work has explored motion-level controllability. ATI~\cite{ati} learns to generate videos from trajectory prompts by interpolating features in latent space, while Go-with-the-Flow~\cite{gowithflow} warps Gaussian noise across timesteps to precisely steer object motion. Several approaches focus on camera motion control, for example, CamI2V~\cite{zheng2024cami2v}, CameraCtrl~\cite{he2024cameractrl}, and CamCo~\cite{xu2024camco} encode viewpoint trajectories for accurate camera dynamics. Other efforts emphasize object-centric guidance~\cite{yin2023dragnuwa,wu2024draganything,tanveer2024motionbridge} or box-level motion cues~\cite{wang2024boximator,yang2024direct}. In contrast, our goal is to enable physical property control rather than explicit motion or trajectory specification.

Physical controllability in video generation has emerged along two main directions. The first integrates explicit physics simulators with generative models to ensure physically grounded behavior. PhysGen~\cite{physgen} couples a 2D physics engine~\cite{Blomqvist_Pymunk_2025} with diffusion-based generation, while WonderPlay~\cite{li2025wonderplay} employs material point methods (MPM)~\cite{mpm} to simulate interactions using Gaussian splats, leveraging diffusion models to enhance photorealism. Similar hybrid approaches~\cite{physmotion,physgaussian,physdreamer,Spring-Gaus,chen2025vid2sim} use MPMs or spring-mass systems to incorporate physical plausibility. PhysDreamer~\cite{physdreamer} inverts physical parameters from generated motion, and PhysAnimator~\cite{PhysAnimator} generates sketch-based physics cues to guide diffusion synthesis. Although these methods yield physically consistent results, they rely on complex simulation pipelines at test time, limiting their scalability and flexibility.

The second direction pursues implicit physical control by embedding physics priors directly into diffusion models without explicit simulators. VLIPP~\cite{vlipp} leverages vision-language models (VLMs) to plan motion trajectories for diffusion guidance, while PhysCtrl~\cite{physctrl2025} introduces a learned point-cloud trajectory generator to control motion within pre-trained diffusion models. While these methods achieve controllability without simulation, they primarily emphasize kinematic guidance. Closest to our work is Force Prompting~\cite{forceprompting}, which fine-tunes a video diffusion model on a 15K-video dataset annotated with force directions. Unlike their single-attribute setup, our method enables control over diverse physical attributes including friction, restitution, external force, and deformation through spatially aligned physical property maps, offering richer and more general forms of physical conditioning.

\vspace{0.2cm}
\noindent \textbf{Reward Optimization for Video Generation.}
Reward-based optimization has become a key paradigm for aligning diffusion models with desired downstream behaviors in both image and video domains. Early methods such as ImageReward \cite{NEURIPS2023_33646ef0} and DFTM \cite{clark2024directlyfinetuningdiffusionmodels} learn differentiable reward functions that enable direct fine-tuning of diffusion models toward human or task-specific preferences. Prabhudesai et al. further advanced this idea to the video domain with VADER \cite{prabhudesai2024videodiffusionalignmentreward,prabhudesai2024aligningtexttoimagediffusionmodels}, introducing gradient-based reward alignment for video diffusion models. More recently, Luo et al. \cite{luo2025dualprocessimagegeneration} employed VLM feedback as differentiable rewards for guiding image generation, and Kumari et al. \cite{kumari2025npedit} demonstrated VLM-based optimization for image editing without paired supervision. In contrast, our work leverages VLM feedback specifically for physics-aware reward optimization, where targeted queries about friction, restitution, deformation, and applied forces guide a controllable video generation model toward physically consistent and interpretable outputs.

%% file: figures/pipeline.tex
\begin{figure*}
    \centering
    \includegraphics[trim={0 8.9cm 0cm 0}, clip, width=0.95\linewidth]{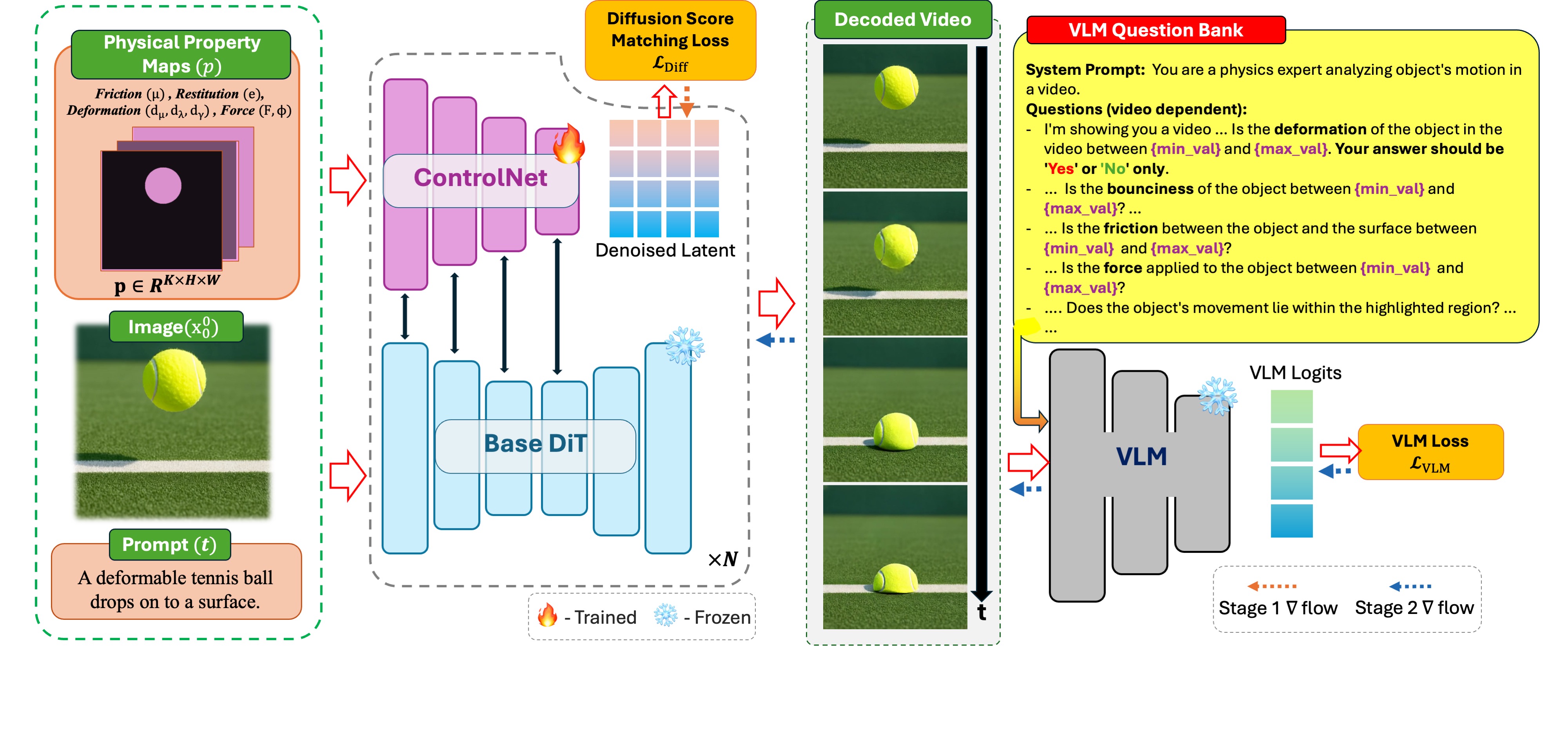}
    \caption{Overview of the proposed PhyCo two-stage training pipeline.
    In Stage 1 (Physics-Supervised Fine-Tuning), the base DiT model is conditioned via a ControlNet using physically rich simulation data, enabling controllable video generation with respect to key physical properties such as friction, restitution, deformation, and applied forces. The model is optimized using a diffusion score matching loss to achieve realistic dynamics. In Stage 2 (VLM-Guided Alignment), the generated videos are analyzed by a frozen Vision-Language Model (VLM) using a curated Physics Question Bank that queries property-specific behaviors (e.g., deformation magnitude, frictional effects, motion alignment). The model receives reward gradients based on VLM logits and responses, guiding the model towards physically plausible and interpretable dynamics.
    }
    \label{fig:pipeline}
\end{figure*}

%% file: tables/dataset_compare2.tex
\begin{table}[t]
\centering
\resizebox{\linewidth}{!}{
\begin{tabular}{lcccccc}
\toprule
\textbf{Dataset \& Benchmarks} &
\textbf{Dataset Size} & 
\textbf{Photo-real} & 
\textbf{Object Dyn.} &
\textbf{Viewpoints} &
\makecell{\textbf{Physical Property}\\\textbf{Annotations}} \\
\midrule
CLEVRER \cite{yi2019clevrer} & 20k videos & \xmark & Multiple & \xmark & \xmark \\
CoPhy \cite{baradel2019cophy} & 238k videos & \xmark & Single & \xmark & \makecell{F, M, G \\ (val only)} \\
ComPhy \cite{chen2022comphy} & 8k videos & \makecell{Partial \\ (syn+real)} & Multiple & \xmark & \makecell{M, C} \\
IntPhys \cite{riochet2018intphys} & 15k videos & \xmark & Single & \xmark & \xmark \\
Physion \cite{bear2021physion} & 16k videos & \xmark & Single & \xmark & \xmark \\
Physion++ \cite{physionpp} & 8k videos & \xmark & Single & \xmark & \makecell{F, M, R, D} \\
ShapeStacks \cite{groth2018shapestacks} & 20k images & \xmark & N/A & \cmark & \makecell{Stability only} \\
Force Prompting \cite{forceprompting} & 38k videos & \cmark & Single & \cmark & Force (implicit) \\
\midrule
\rowcolor{blue!8}
\textbf{PhyCo (Ours)} & \textbf{100k videos} & \textbf{\cmark} & \textbf{Multiple} & \textbf{\cmark} & \makecell{\textbf{F, M, R, D} \\ \textbf{Force}} \\
\bottomrule
\end{tabular}}
\caption{Comparison of physics-rich datasets and benchmarks used for learning and evaluating object dynamics. F, M, R, D, and G denote friction, mass, restitution, deformation, and gravity, respectively.}
\label{tab:dataset_comparison}
\end{table}

%% file: sec/3_1_simdata.tex
\section{Method}

Our goal in this work is to enable diffusion models with continuous and interpretable control over key physical properties—friction, restitution, deformation, and applied forces, while maintaining photorealistic synthesis and broad generalization. Achieving this requires both (i) a source of physically meaningful supervision that fairly generalizes beyond the training domain and (ii) a mechanism to inject such information into a pretrained generative model. We therefore introduce a unified pipeline that couples physically grounded simulation data with a two-stage training procedure: physics-supervised fine-tuning and VLM-guided reward optimization.


\subsection{Physically Grounded Simulations}

Although producing synthetic data with physics engines is straightforward, creating simulations that meaningfully benefit controllable video generation is far from trivial. The challenge is not the quantity of data but the quality and relevance of the physical behaviors it expresses. In practice, we find that simulations must satisfy two criteria to effectively teach controllable physics priors:
(1) the physical attribute of interest must manifest clearly and unambiguously in the visual motion, and
(2) the scenario must lie within the competence range of the pretrained diffusion backbone.
Overly complex scenes—such as interactions involving multiple objects or cluttered dynamics—often produce outputs that current diffusion models struggle with, as also noted in recent studies such as PISA \cite{li2025pisaexperimentsexploringphysics}. Training on such data introduces unnecessary variance and slows learning. In contrast, focused, well-structured simulations accelerate controllability by letting the model associate each physical property with its canonical visual signature, similar to the design philosophy behind Force-Prompting \cite{forceprompting}.

Following this insight, we construct a large suite of physics-rich but visually clean simulation scenarios using the Kubric framework \cite{greff2021kubric, coumans2016pybullet, blender}. Each simulation isolates or combines fundamental object–environment interactions while systematically varying four key physical parameters—friction, restitution, deformation, and applied force. These variations allow the model to observe consistent motion changes tied directly to interpretable physical attributes.

Our dataset spans six controlled scenarios designed to expose different dynamic regimes: a brick sliding on a flat plane, a ball rebounding off a wall, a vertically bouncing ball, a soft ball dropping under gravity, an object impacting a  deformable body, and multiple balls colliding on a pool table. Each sequence is randomized in object color, surface material, camera placement, and HDRI illumination (50 environments following \cite{forceprompting}), with high-quality textures from Polyhaven to enhance photorealism. This balance of controlled physics and appearance diversity helps the diffusion model disentangle visual variation from underlying dynamics. In total, we generate more than 100K videos across all scenes. The resulting dataset provides a structured yet diverse foundation for learning controllable, interpretable, and physically grounded video dynamics.

\input{figures/sim_videos}

%% file: figures/sim_videos.tex
\begin{figure}
    \centering
    \includegraphics[trim={0 32cm 33cm 0}, clip, width=1.0\linewidth]{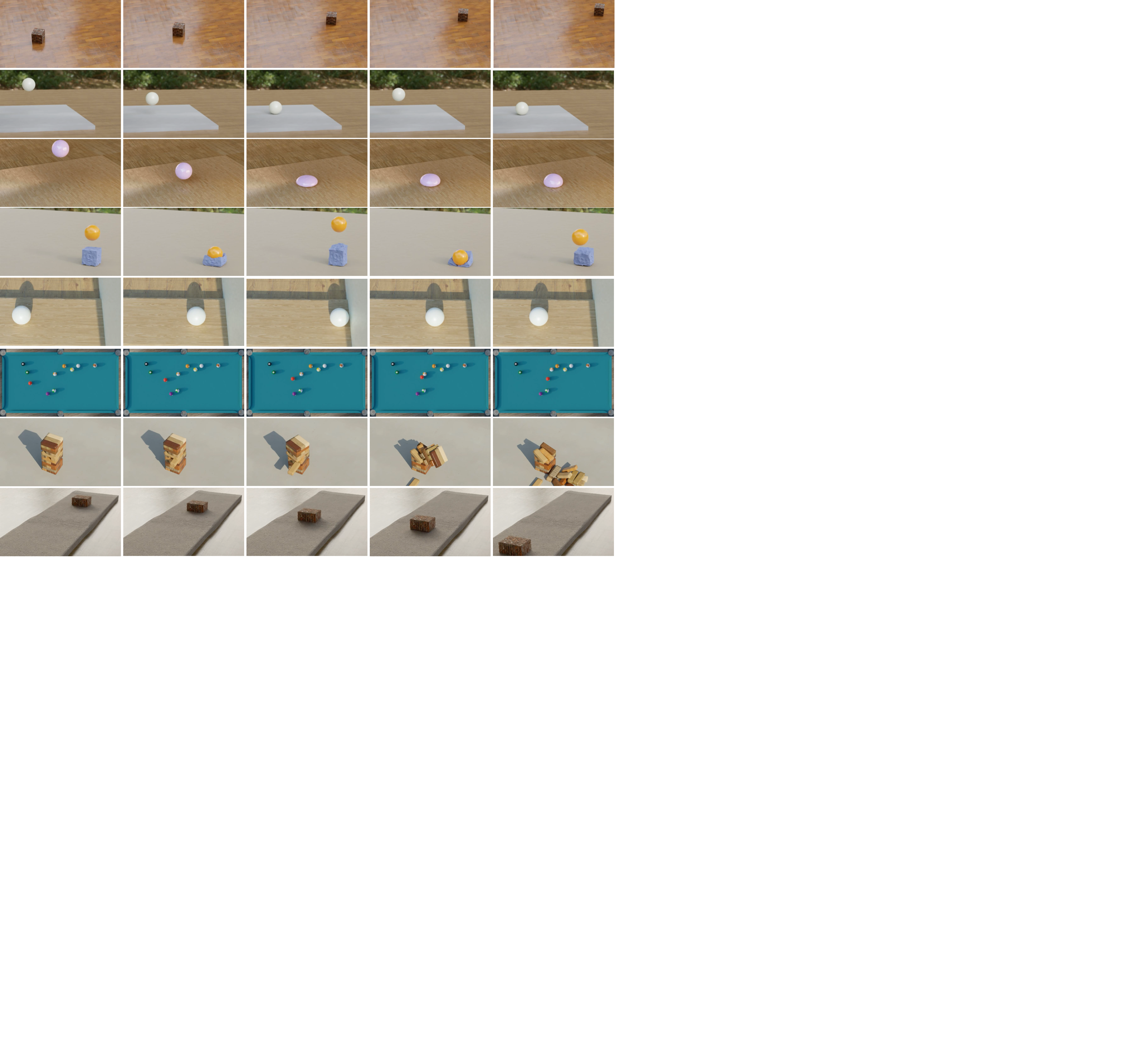}
    \caption{Example simulation videos from our dataset comprising of more than 100K videos from 8 different scenarios.}
    \label{fig:sim_videos}
\end{figure}

%% file: sec/3_2_method.tex
\subsection{Physics Supervised Fine-tuning}
The overall pipeline is illustrated in Fig.~\ref{fig:pipeline}. We fine-tune a pre-trained Cosmos-Predict2-2B~\cite{nvidia2025cosmosworldfoundationmodel} diffusion backbone by conditioning the denoising process on physical property inputs using a ControlNet-style~\cite{controlnet} architecture. Formally, the generator $G_\theta$ models the conditional distribution $p_\theta(\mathbf{x}_{1:T} \vert \mathbf{t}, \mathbf{x}_0^0, \mathbf{p})$, where $\mathbf{x}_{1:T}$ is the target video sequence, $\mathbf{t}$ the text prompt, $\mathbf{x}_0^0 \in \mathbb{R}^{C \times H \times W}$ the initial frame, and $\mathbf{p} \in \mathbb{R}^{K \times H \times W}$ encodes spatially aligned physical attributes.


\noindent \textbf{Encoding Physical Property Maps.} 
To enhance compactness and promote better generalization from limited training data, we represent objects as spatially aligned circular blobs. The ControlNet receives a tokenized representation of the property field $\mathbf{p}$ where each attribute is normalized between $[-1, 1]$. We decompose $\mathbf{p}$ into groups $\{\mathbf{p}^{(g)}\}_{g=1}^{G}$, where each $\mathbf{p}^{(g)} \in \mathbb{R}^{3 \times H \times W}$ contains semantically related channels: (1) friction $\mu_f$ and restitution $e$, padded with a constant channel; (2) Neo-Hookean deformation parameters $d_\mu$, $d_\lambda$, $d_\gamma$; and (3) force magnitude $F$ and direction represented by $(\cos\phi, \sin\phi)$. Each group $\mathbf{p}^{(g)}$ is tokenized by the Cosmos tokenizer $\tau(\cdot)$, producing embeddings $\mathbf{z}^{(g)} = \tau(\mathbf{p}^{(g)}) \in \mathbb{R}^{L_g \times D},$
which are linearly projected to form the conditioning sequence. 

\noindent \textbf{Training.} We finetune only the ControlNet layers while keeping the base diffusion model and tokenizer weights frozen to preserve their pretrained representations. The encoded physical property maps $\mathbf{z}^{(g)}$ are first passed through an adapter network $A(\cdot)$ that projects the tokenized property embeddings $\mathbf{h}_p = A(\mathbf{z}_p)$ to match the input dimensionality of the DiT backbone. Each semantic group $\mathbf{h}_p^{(g)}$ is processed by a separate ControlNet branch to enable faster training and compositionality of various physical properties. Each ControlNet is trained on a subset of PhyCo dataset where the corresponding input physical property manifests in the observed dynamics. The optimization follows the diffusion score-matching objective used in the Cosmos World Foundation Model~\cite{nvidia2025cosmosworldfoundationmodel,diffusionLoss1,diffusionLoss2}, ensuring consistent noise scheduling and temporal supervision across frames.

\subsection{VLM Reward Optimization}
The second stage of our training leverages feedback from a foundational Vision–Language Model (VLM) to refine both the controllability and physical alignment of the generated videos. While supervised fine-tuning on physics-rich simulation datasets yields visually coherent and physically plausible results, it alone does not guarantee strong control fidelity. To bridge this gap, we employ a VLM as a generalized critique model that evaluates physical consistency through targeted, physics-aware queries. The VLM outputs token logits that are then converted into a differentiable reward signal, guiding the generator toward physically interpretable and controllable behaviors.

\noindent \textbf{Approach.}
Standard diffusion training with score matching involves denoising a noised version of the ground-truth video. However, such single-step reconstructions are unsuitable for VLM evaluation due to two main reasons: (i) the visual details and object boundaries remain blurry, and (ii) they already encode the global physical trajectory from the conditioning signal (e.g., applied force direction), which masks the true inference-time behavior of the model. For instance, a partially noised ground-truth video of an object under an applied force still shows its dominant motion direction, even though the model might fail to reproduce it at inference time.

To obtain a faithful proxy of the inference-time generation process, we instead perform an $N$-step denoising rollout to generate a predicted latent $\hat{\mathbf{z}}_0$ given the initial frame $\mathbf{x}_0^0$, text prompt $\mathbf{t}$, and physical property maps $\mathbf{p}$. The latent is decoded to a video $\hat{\mathbf{x}}_0 \in \mathbb{R}^{T \times C \times H \times W}$, which serves as the input to the VLM along with a structured set of physics queries. Our formulation is inspired by recent VLM-based optimization works such as \cite{luo2025dualprocessimagegeneration, kumari2025npedit}, but differs in that our feedback focuses on physical controllability of videos rather than semantic or aesthetic alignment for image editing.

\noindent \textbf{VLMs for Physical Question Answering.}
While off-the-shelf VLMs excel at recognizing explicit visual cues, they often struggle with implicit physical reasoning such as motion consistency, restitution, or frictional response. To improve reliability, we fine-tune Qwen2.5-VL-3B for 200 steps using our synthetic simulation dataset, where each clip is paired with multiple physics-related questions (e.g., “Does the object move in the intended direction of force?”). This short adaptation yields high accuracy ($\approx85\%$) within 100 iterations, enabling robust reward computation. 

\input{tables/physics_iq}

\noindent \textbf{Physical Alignment Reward.}
We structure each query as a binary (``Yes'' / ``No'') question following prior work in vision-language reward learning \cite{luo2025dualprocessimagegeneration,kumari2025npedit}. For each generated video $\hat{\mathbf{x}}_0$, we compute a reward by comparing VLM logits corresponding to the correct and incorrect answer tokens, guided by ground-truth property ranges $\mathbf{p} \in [0,1]$. Each physical attribute (e.g., friction, deformation, restitution, force magnitude or direction) is probed with multiple thresholded questions over $\{\textit{min\_val}, \textit{max\_val}\}$ to obtain dense feedback signals. To assess directional consistency, we overlay a blue angular sector on the video and query whether the motion lies within this region.

We compute the VLM alignment loss as a binary cross-entropy over the logit difference between correct and incorrect answer tokens:
\begin{equation}
\mathcal{L}_{\text{VLM}} = - \sum_i \log \sigma(\zeta_+^{(i)} - \zeta_-^{(i)}),
\end{equation}
where $\zeta_+^{(i)}$ and $\zeta_-^{(i)}$ are the logits for the correct and incorrect responses to the $i$-th question, respectively.

\noindent \textbf{Training.}
We fine-tune only the ControlNet layers of our diffusion model corresponding to the physical property maps $\mathbf{p}$ using the VLM-based reward loss $\mathcal{L}_{\textnormal{VLM}}$, excluding the diffusion score-matching objective. This focused optimization yields more stable and physically consistent generations compared to joint training with score matching. 
For meaningful feedback, we perform a 10-step denoising rollout, decode the latent into a video, and backpropagate the VLM loss end-to-end through the VLM, tokenizer, and DiT backbone.

%% file: tables/physics_iq.tex
\begin{table*}[t]
    \centering
    \resizebox{1.0\linewidth}{!}{
    \begin{tabular}{lcccccc}
        \toprule
        \textbf{Model} & \textbf{Solid Mechanics ($\uparrow$)} & \textbf{Fluid Dynamics ($\uparrow$)} & \textbf{Optics ($\uparrow$)} & \textbf{Magnetism ($\uparrow$)} & \textbf{Thermodynamics ($\uparrow$)} & \textbf{IQ Score ($\uparrow$)} \\
        \midrule
        SVD-XT~\cite{blattmann2023stablevideodiffusionscaling}          & 21.9 & 20.5 & 6.8  & 8.4  & 17.1 & 19.1 \\
        LTX-Video-I2V~\cite{hacohen2024ltxvideorealtimevideolatent}   & 30.2 & 29.8 & 15.9 & 13.2 & 8.4  & 26.8 \\
        SG-I2V~\cite{namekata2025sgi2vselfguidedtrajectorycontrol}         & 34.6 & 31.2 & 15.9 & 13.1 & 8.4  & 29.7 \\
        Cogvideo-I2V-5B~\cite{hong2022cogvideo} & 30.4 & 29.8 & 16.7 & 13.3 & 8.5 & 27.1 \\
        Cosmos-Predict2-2B \cite{nvidia2025cosmosworldfoundationmodel} & 31.7 & 25.2 & 26.2 & 9.1 & 16.9 & 27.7 \\
        VLIPP\cite{vlipp}            & 42.3 & 34.1 & 16.9 & 13.4 & 8.8  & 34.6 \\
        \midrule
        \multicolumn{7}{c}{Test time extrapolated generation: 120 frames @ 24FPS} \\
        \midrule
        Ours (Text only) & 36.5  & 28.9 & 18.9 & 12.6 & 32.0 & 30.9\\
        Ours (ControlNet) & 42.3 & 30.7 & 19.3 & 12.6 & 40.1 & 35.3\\
        Ours (ControlNet + VLM Loss) & 44.1 & 31.2 & 20.1 & 17.2 & 33.1 & 36.3\\
        \midrule
        \multicolumn{7}{c}{Train-time conditions: 57 frames @ 24FPS + last-frame repetition} \\
        \midrule
        Ours (Text only) & 43.9  & 38.5 & 17.5 & 21.7 & 26.8 & 36.5\\
        Ours (ControlNet) & 49.7 & 37.8 & 16.3 & 19.9 & 18.2 & 38.9 \\
        Ours (ControlNet + VLM Loss) & 53.1 & 44.3 & 20.3 & 20.8 & 35.9 & 43.6 \\
        \bottomrule
    \end{tabular}
    }
    \caption{Quantitative results of physically plausible video generation on Physics-IQ Benchmark.}
    \label{tab:physics_iq_results}
\end{table*}

%% file: sec/4_results.tex
\input{figures/controllable_results}
\section{Experimental Results}

\input{figures/composition_results}

In this section, we evaluate our approach across quantitative metrics, qualitative comparisons, user preferences, and ablation studies. Our experiments aim to answer four core questions about our method: (i) generating plausible physical dynamics,  (ii) fine-grained controllability over physical attributes (iii) VLM reward optimization to enhance physical fidelity and (iv) generalization across scenarios and new objects. To faithfully answer these questions we compare our method against several state-of-the art video generation models across physics benchmarks and other carefully designed controlled experiments and studies.

\noindent \textbf{Baselines.} Our primary baselines include text-conditioned image-to-video world models such as Cosmos-Predict2~\cite{nvidia2025cosmosworldfoundationmodel}, CogVideoX-I2V-5B~\cite{yang2024cogvideox}, SVD-XT~\cite{blattmann2023stablevideodiffusionscaling} and LTX-Video-I2V~\cite{hacohen2024ltxvideorealtimevideolatent}. We also evaluate a text-only fine-tuned variant of Cosmos-Predict2 trained on our proposed PhyCo dataset, to measure the benefit of physics-aware supervision. We further compare against two baselines closest to our approach, Force-Prompting~\cite{forceprompting} and VLIPP~\cite{vlipp}. Force-Prompting improves controllability and physics awareness through force-specific supervision, whereas VLIPP uses a VLM to extract coarse motion trajectories (with bounding boxes) that guide the diffusion model toward physically consistent outputs.

\input{tables/user_study}
\input{tables/vlm_loss_table}

\input{figures/baselines_fp}
\input{figures/ablations_fig}

\noindent \textbf{Quantitative Evaluation on Physics-IQ Benchmark.} We first evaluate our method on the Physics-IQ \cite{physicsiq} benchmark, which measures the physical realism of generated videos across five domains—Solid Mechanics, Fluid Dynamics, Optics, Magnetism, and Thermodynamics. The benchmark computes a \emph{Physics-IQ} score by comparing the timing and spatial alignment of key actions in generated videos against real-world reference sequences (396 videos). Results on this benchmark are presented in Tab.~\ref{tab:physics_iq_results}. 
While the Physics-IQ benchmark evaluates 5-second videos (120 frames at 24 FPS), our model is trained on 57-frame sequences. Nevertheless, even under this train and test time mismatch, our model significantly outperforms several state-of-the-art open-source video generation systems across all evaluated categories. For completeness, we additionally report results obtained when we adhere to our training-time conditions by generating 57 frames and repeating the final frame to match the benchmark duration. 
Both these results paint a consistent picture: our physics-aware conditioning leads to more realistic and physically coherent dynamics, underscoring the benefit of the proposed PhyCo dataset and the robustness of our approach even when tested beyond the training domain.

\noindent \textbf{User Study.}
We conduct a 2AFC study with 16 participants, each comparing 39 video pairs differing in a single physical attribute. Across 98 generated videos per method, users consistently prefer PhyCo in Table \ref{tab:user_study}, indicating more realistic physical behavior and clearer variations. This confirms that our conditioning improves both controllability and perceived physical plausibility.
\noindent \textbf{Ablations on PhyCo Data.}
To quantitatively assess how well generated videos adhere to the input physical properties, we conduct evaluations on an in-domain simulation test set consisting of 100 videos spanning all attributes. 
A fine-tuned Qwen2.5-VL-3B model~\cite{qwen2025qwen25technicalreport} predicts physical properties from generated outputs, which we compare to the ground-truth conditioning inputs.
Results of this analysis are presented in Tab.~\ref{tab:ablation_study}. As shown, models trained with explicit VLM-based reward optimization achieves significantly better alignment with the intended input properties, confirming that our reward formulation strengthens controllability and ensures more faithful adherence to physical conditioning during generation (Fig.~\ref{fig:ablation_fig}).

\noindent \textbf{Force Direction Adherence.}
We assess force-conditioned controllability on 25 real-world videos by applying random force directions and measuring the angular deviation between intended and observed motion. Our model achieves a substantially lower mean directional error ($15.2^\circ$) compared to Force-Prompting ($40.5^\circ$), indicating more reliable and precise control over induced dynamics. As shown in Fig.~\ref{fig:baselines_w_fp}, Force-Prompting fails to produce the desired motion—particularly in scenes where large or visually dominant objects are present, while our method consistently generates motion in the correct direction.

\noindent \textbf{Qualitative Results.} Representative snapshots highlighting controllability and compositionality over several physical attributes is shown in Figures ~\ref{fig:teaser}, ~\ref{fig:results_fig} and ~\ref{fig:compositionality_fig} respectively. 
Despite being trained solely on synthetic simulations, the model generalizes effectively to new object categories, motion types and appearance shifts (Fig.~\ref{fig:teaser}), exhibiting stable and coherent behavior under varying physical properties. For example, a model trained only on simple bouncing-ball simulations can generalize to a person jumping on a trampoline—where low restitution settings result in no rebound after impact. Likewise, training on simple blocks sliding across flat surfaces extends naturally to more complex object and surface configurations. These results demonstrate the strong generalization capabilities of our approach compared to the baselines (Fig.~\ref{fig:baselines_w_fp}) and underscores the value of our simulated dataset in enabling physically consistent and controllable video generation.

%% file: figures/controllable_results.tex
\begin{figure*}
    \includegraphics[trim={0 24.5cm 9.5cm 0}, clip, width=1.0\linewidth]{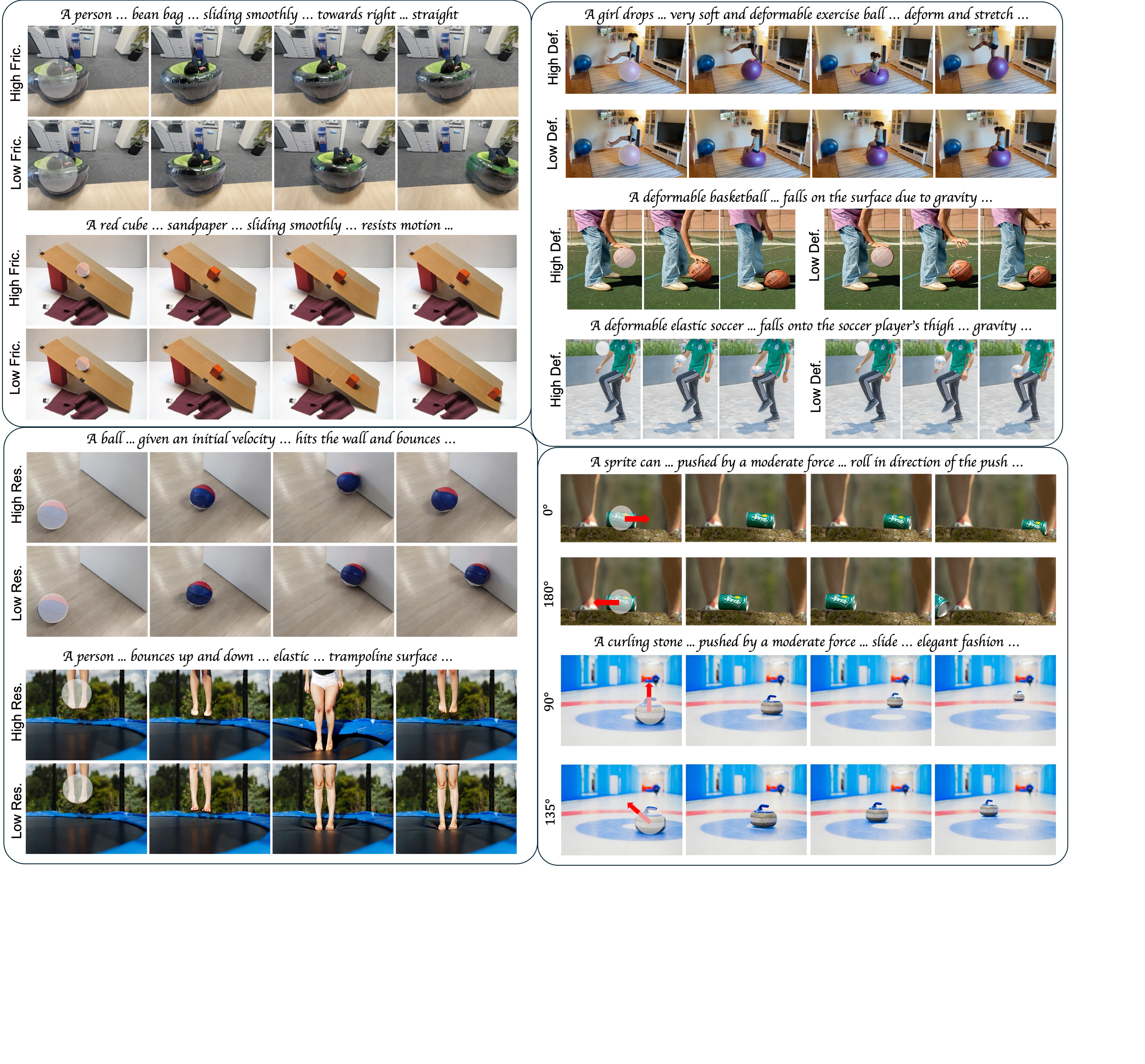}
    \caption{Shows representative frames from our generated outputs, illustrating controllable behavior across key physical attributes including friction, restitution, deformation, and external forces. White blobs in the figure highlights the locations of spatially aligned pixel property inputs.
    These results highlights the ability of our model to generalize beyond the training domain for controllable physically consistent generation. For instance, the girl hopping on the exercise ball exhibits a noticeably higher bounce when the ball’s deformability is increased, compared to the lower-deformation setting.
    }
    \label{fig:results_fig}
\end{figure*}

%% file: figures/composition_results.tex
\begin{figure*}[h]
    \includegraphics[trim={0 85cm 9.5cm 0}, clip, width=1.0\linewidth]{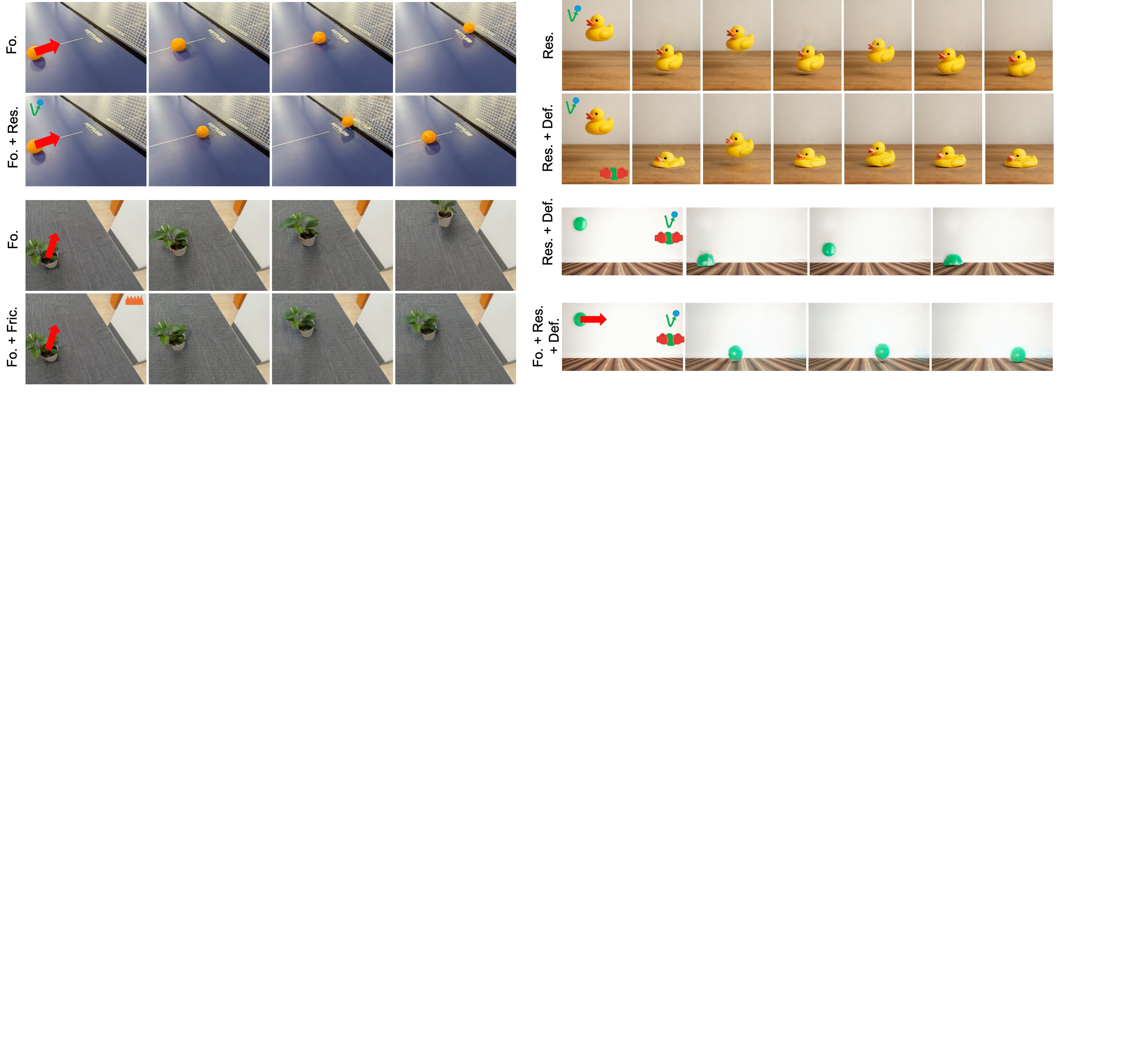}
    \caption{Qualitative results illustrating compositional control over multiple physical attributes within the same scene. Force, friction, restitution, and deformation are denoted as Fo, Fric, Res, and Def, respectively.}
    \label{fig:compositionality_fig}
\end{figure*}

%% file: tables/user_study.tex
\begin{table}[ht]
    \centering
    \resizebox{\linewidth}{!}{
    \begin{tabular}{c|c|c|c|c}
        \toprule
        Ours vs. Methods & Friction & Restitution & Deformation & Force \\
        \midrule
        Force Prompting \cite{forceprompting} & -- & -- & -- & 71.7\% \\
        CogVideoX-I2V-5B~\cite{yang2024cogvideox} & 95.5\% & 100.0\% & 82.2\% & 91.1\% \\
        Cosmos-Predict2B~\cite{nvidia2025cosmosworldfoundationmodel} & 100.0\% & 93.2\% & 91.3\% & 86.4\% \\
        Ours (Text only) & 90.9\% & 67.4\% & 56.8\% & 58.7\% \\
        \bottomrule
    \end{tabular}
    }
    \caption{User study results evaluated on the physical realism axis for different physical properties. Physical realism reports the percentage of pairwise preferences (\%) for our method over the baselines in a 2AFC human evaluation, where scores above 50\% indicate a preference for our ControlNet generations.}
    \label{tab:user_study}
\end{table}

%% file: tables/vlm_loss_table.tex
\begin{table}[t]
\centering
\scriptsize
\resizebox{\linewidth}{!}{
\begin{tabular}{l|ccccc}
\toprule
Method & FM & Fric. & FD ($^\circ$) & Res. & Def. \\
\midrule
Base Model (zero-shot) \cite{nvidia2025cosmosworldfoundationmodel} & 0.38 & 0.33 & 91.87 & 0.40 & 0.45 \\
Text-only (finetuned) & 0.31 & 0.30 & 40.35 & 0.31 & 0.14 \\
ControlNet (–VLM) & 0.33 & 0.24 & 38.05 & 0.28 & 0.14 \\
ControlNet (+VLM) & \textbf{0.28} & \textbf{0.20} & \textbf{22.53} & \textbf{0.16} & \textbf{0.10} \\
\bottomrule
\end{tabular}
}
\caption{Ablation study on synthetic data across five controllable properties: 
FM = force magnitude error, Fric. = friction error, FD = angular deviation in force direction (lower is better), Res. = restitution error, and Def. = deformation error.}
\label{tab:ablation_study}
\end{table}

%% file: figures/baselines_fp.tex
\begin{figure}
    \centering
    \includegraphics[trim={0 60cm 64cm 0}, clip, width=1.0\linewidth]{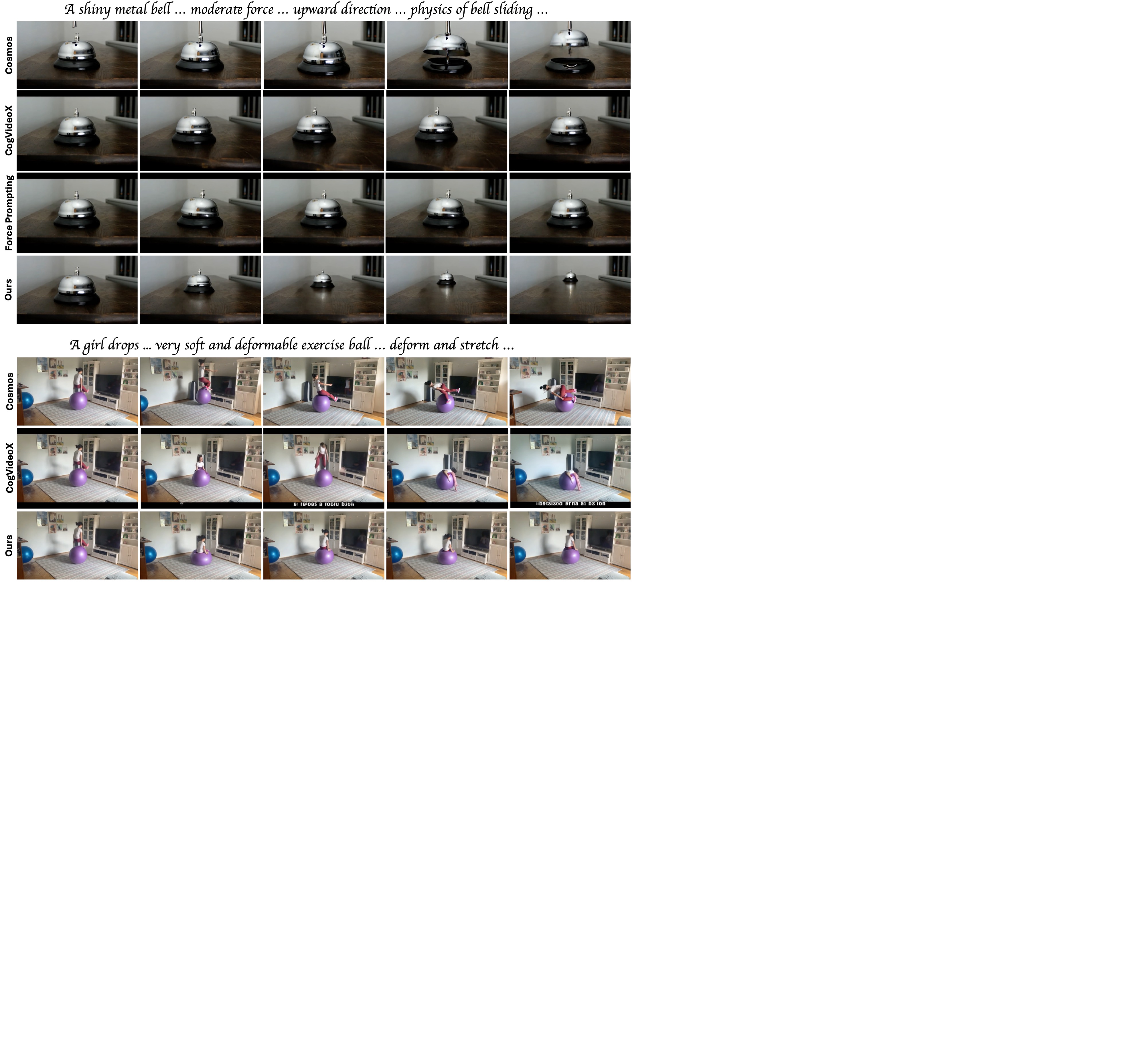}
    \caption{Qualitative comparison between our proposed method and other baselines. Note, without force inputs, Force-Prompting~\cite{forceprompting} is same as CogVideoX~\cite{yang2024cogvideox} model.}
    \label{fig:baselines_w_fp}
\end{figure}

%% file: figures/ablations_fig.tex
\begin{figure}
    \centering
    \includegraphics[trim={0 94.5cm 64cm 0}, clip, width=1.0\linewidth]{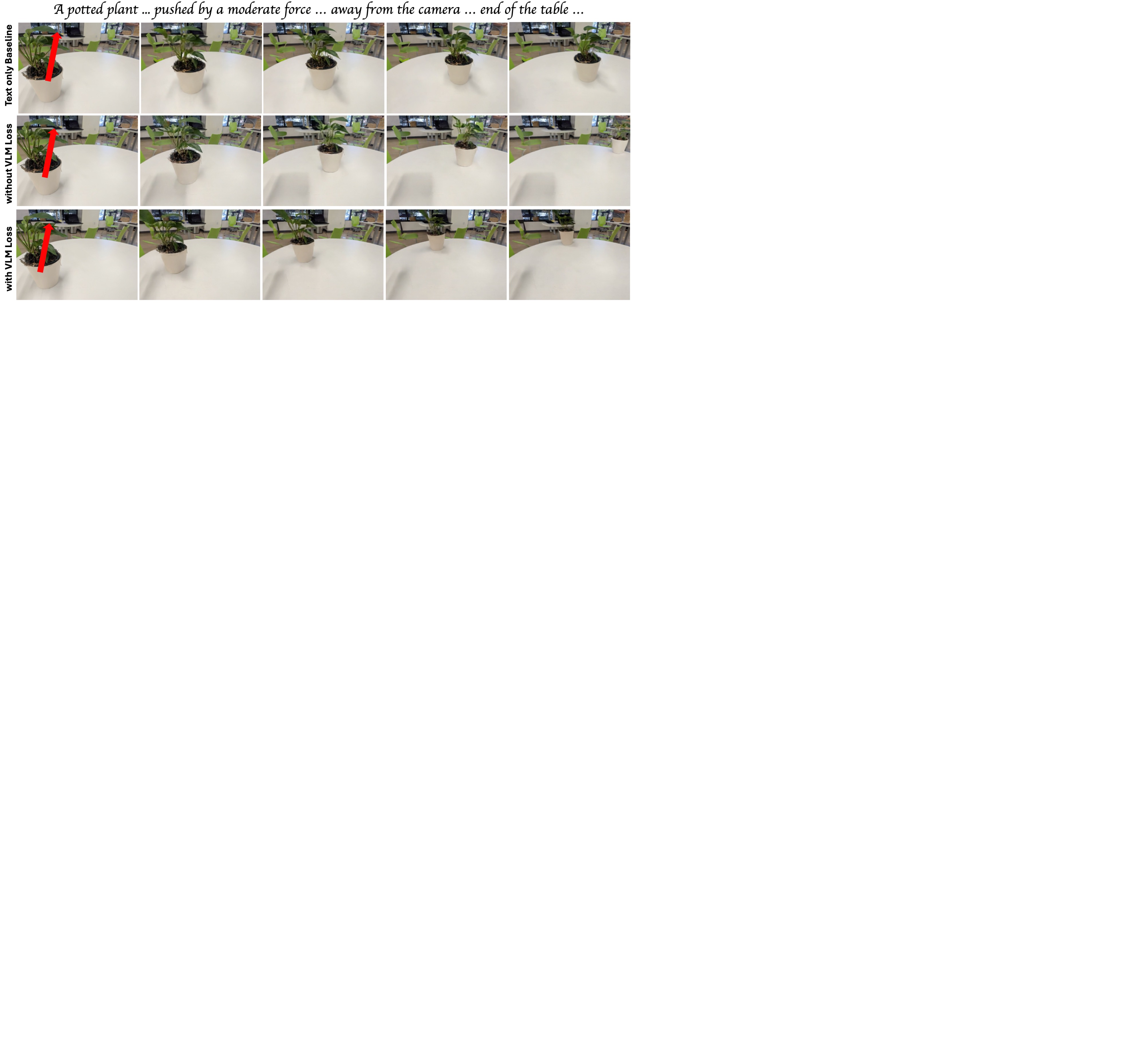}
    \caption{Ablation on explicit physical property conditioning using a ControlNet with additional VLM feedback.}
    \label{fig:ablation_fig}
\end{figure}

%% file: sec/5_conclusion.tex
\section{Conclusion}
We presented PhyCo, a controllable video generation framework that injects physically grounded priors into diffusion models through explicit property conditioning and VLM-guided reward optimization. Our approach enables continuous control over key dynamics such as friction, restitution, deformation and force, without requiring simulators at inference. Extensive evaluations show clear gains in both physical realism and controllability, with strong generalization beyond synthetic training data. 
These results point to a scalable path toward physically consistent, controllable video models that generalize reliably to real-world dynamics.

%% file: sec/X_suppl.tex
\clearpage
\appendix
\setcounter{page}{1}
\maketitlesupplementary

\section{Video Results on Webpage}
All video results are available at \href{https://phyco-video.github.io/}{\texttt{phyco-video.github.io}}. The webpage includes both the examples shown in the main paper and additional results not included due to space constraints. We organize the content as follows:

\vspace{0.15cm}
\noindent \textbf{Generalization Across Artistic Styles.}
We present multi-style storylines where four key frames guide the model to generate physically consistent sequences. Figure~\ref{fig:storyline_ff} shows the first frames used in these examples.

\input{figures/ff_different_styles}

\vspace{0.15cm}
\noindent \textbf{Fine-Grained Control of Physical Properties.}
We provide videos generated under three distinct levels (low, medium, high) for each physical attribute to demonstrate smooth and continuous control.

\vspace{0.15cm}
\noindent \textbf{Compositionality of Multiple Attributes.}
We showcase combinations such as force+friction, force+bounciness, and bounciness+deformation. Even though some parameter pairings (e.g., restitution and deformation) are difficult to simulate accurately in current physics engines~\cite{coumans2016pybullet}, our model still achieves visually convincing behavior.

\vspace{0.15cm}
\noindent \textbf{Baseline Comparisons.}
Side-by-side videos compare our approach with recent video diffusion models. We also include interactive control over force direction with pre-generated results.

\section{PhyCo Dataset Details}

The PhyCo dataset consists of physically simulated scenes in PyBullet and photorealistic rendering in Blender. Each sample is a 4-second video (24 FPS) at $432\times768$ resolution. Alongside RGB frames, we provide synchronized depth maps, per-frame segmentation masks, and structured metadata describing scene geometry, object material properties, and all applied physical parameters. We also include a standardized scene-level text description for every video to support multi-modal supervision and VLM-based evaluation.

\vspace{0.2cm}
\noindent \textbf{Sampling Physical Properties.} We vary the key physical parameters that govern object dynamics and deformation. For rigid bodies, friction and restitution coefficients are uniformly sampled between 0 and 1 in PyBullet \cite{coumans2016pybullet}, covering behaviors from smooth sliding to high resistance, and from inelastic impacts to highly bouncy interactions.

For deformable bodies, we use a FEM-based Neo-Hookean model, varying the Lamé coefficients $\mu$ and $\lambda$ along with a damping coefficient $\gamma$ to span materials from nearly rigid to highly deformable. We also simulate a range of external forces, where low to high magnitudes map to gentle interactions and strong impacts. These forces are projected from 3D world coordinates onto the rendered 2D frame using the camera parameters, enabling direct correspondence between physical actions and visual outcomes.

\section{Implementation Details}
This section summarizes implementation specifics for the proposed PhyCo model, including ControlNet fine-tuning and VLM-based supervision, as well as Qwen2.5-VL fine-tuning used for evaluation.

\subsection{PhyCo Implementation Details}

\noindent \textbf{ControlNet Training.}
We fine-tune only the ControlNet layers, while keeping the base video diffusion model and tokenizer weights frozen to preserve the pretrained dynamics learned by the Cosmos World Foundation Model~\cite{nvidia2025cosmosworldfoundationmodel}. Training is performed using 4×H100 GPUs for 10k optimization steps (approximately half a day) per attribute-specific ControlNet branch. We supervise 57 frames per sequence at 24 FPS, employing a per-device batch size of 1 with 2 steps of gradient accumulation, yielding an effective batch size of 8. We use a learning rate of $2^{-14.5}$ and maintain a peak memory footprint of approximately 45 GB per GPU. Optimization follows a standard diffusion score-matching loss~\cite{diffusionLoss1,diffusionLoss2} with consistent noise scheduling and temporal supervision.

\vspace{0.2cm}
\noindent \textbf{VLM-Based Reward Optimization.}
To incorporate physics-aware perceptual supervision, each ControlNet branch is further trained with a VLM-guided reward loss for 100 iterations (roughly 70 minutes). Videos are spatially downsampled to half resolution and temporally subsampled to a maximum of 16 frames before being fed to the VLM. This configuration uses 8×H200 GPUs with an effective batch size of 4 and requires up to 115 GB VRAM. Reducing the number of generated frames or directly aligning DiT latents with the VLM input would further lower memory needs—an avenue we leave for future work.

\vspace{0.2cm}
\subsection{Qwen2.5-VL Fine-tuning for Evaluation}
We adapt Qwen2.5-VL-3B~\cite{qwen2025qwen25technicalreport} on the PhyCo dataset to robustly infer physical properties from video inputs. Training samples are generated using the physics-focused queries listed in Fig.~\ref{vlm_prompts_box}. For all binary (Yes/No) questions, we ensure a balanced set of responses. Videos are temporally subsampled to at most 16 frames. For force-direction queries, we add a highlighted blue sector overlay indicating the target force angle (see Fig.~\ref{fig:rollout_overlay}). We fine-tune for 200 iterations using LoRA with rank and $\alpha$ set to 64, across 4×H100 GPUs with an effective batch size of 128 and learning rate of $2\times10^{-4}$.

\section{Additional Results}

\noindent \textbf{Motion Consistency Evaluation.}
We further evaluate motion consistency using the Fréchet Video Motion Distance (FVMD)~\cite{liu2024frechetvideomotiondistance} on Physics-IQ benchmark videos (Table~\ref{tab:fvmd_scores}). FVMD measures the distributional distance between generated and reference motion features, capturing temporal dynamics independently of appearance quality, with lower values indicating more realistic motion.

Our method achieves the best or second-best FVMD scores across most domains, demonstrating improved temporal coherence and physically plausible motion. Notably, incorporating VLM-based reward optimization consistently yields further gains over ControlNet without VLM, particularly in solid mechanics, fluid dynamics, and magnetism. These trends closely mirror the Physics-IQ results, indicating strong alignment between perceptual physics reasoning and motion statistics. Minor deviations are observed in thermodynamics, likely due to the limited number of evaluation scenes in this domain.


\input{tables/fvmd_scores}


\noindent \textbf{Generalization Across Backbones.}
To demonstrate that our dataset enables physically consistent generation beyond a specific architecture, we finetune a Wan2.2 video model using only text conditioning. As shown in Table~\ref{tab:wan22_physics_iq}, finetuning the Wan2.2 base model on the proposed PhyCo dataset yields a 
4.6$\%$ improvement in average on Physics-IQ score. This gain highlights the effectiveness of PhyCo in imparting physically meaningful priors, even without explicit conditioning mechanisms such as ControlNet. Figure~\ref{fig:wan22_rebuttal} further illustrates qualitative examples, where the finetuned Wan2.2 model exhibits controllable physical behavior.

\input{figures/physics_iq_qual}
\input{figures/wan22_rebuttal}
\input{tables/wan22_physics_iq}
\noindent \textbf{Qualitative Results on Physics-IQ Dataset.}
Figure~\ref{fig:physics_iq_qual} presents qualitative comparisons from the Physics-IQ~\cite{physicsiq} benchmark, demonstrating that our method produces dynamics more consistent with real-world physical behavior. For instance, when a ball momentarily occludes behind a bag, it reappears with a coherent trajectory and speed, reflecting improved temporal consistency in motion prediction. Likewise, scenes with contact-induced deformation show physically plausible responses—such as a pillow compressing noticeably under the weight of a kettlebell while remaining largely unaffected by a lightweight paper object. These examples highlight how explicit physical conditioning leads to more realistic and physically interpretable video synthesis across diverse scenarios.

\vspace{0.2cm}
\noindent \textbf{Results from VLM fine-tuning.} We test the ability of the fine-tuned VLM to predict intrinsic physical physical property values from videos by testing it on a held out PhyCo test set of 100 samples. We find the mean absolute error across all four attributes (friction, restitution, deformation and force) to be $0.14$ . Further, the accuracy of prediction for binary responses to be 84.8\% across all four attributes. 

\vspace{0.2cm}
\noindent \textbf{Analysis of Flickering Artifacts.}
We observe that flickering primarily arises in regions with rapid per-frame motion, especially for thin or high-frequency structures. Increasing the training frame rate significantly mitigates these artifacts (Fig.~\ref{fig:fps_ablation_qual}), as it provides denser temporal supervision and reduces abrupt motion discontinuities. In addition, stronger video diffusion backbones such as Wan2.2 exhibit markedly reduced flickering (Fig.~\ref{fig:wan22_rebuttal}), suggesting that improved temporal modeling further enhances stability.

These observations are consistent with our quantitative results: improvements in motion quality are reflected in lower FVMD scores (Tab.~\ref{tab:fvmd_scores}), indicating better temporal coherence. Overall, both higher-FPS training and advances in backbone architectures play a complementary role in reducing flickering and improving motion consistency.

\input{figures/fps_ablation_qual}

\vspace{0.2cm}
\noindent \textbf{Details on User-study.} Illustration of the interface used in conducting the user-study is shown in Fig.~\ref{fig:user_study}
\input{figures/user_study_fig}

\input{figures/force_dir_overlay}

\section{Limitations}
Although our approach improves controllability and physical consistency over existing video diffusion models, the generated dynamics are still an approximation of real physics rather than an accurate reproduction. Our physical priors primarily capture simplified rigid and soft-body behaviors in controlled settings, and more complex interactions—such as articulated motion, fluid-structure coupling, or multi-contact dynamics—remain partially modeled. Additionally, while spatial property maps provide interpretable control, they do not enforce strict adherence to underlying conservation laws (e.g., momentum, deformation energy), occasionally producing subtle but noticeable physical deviations. Extending our framework toward richer physical regimes, stronger real-world grounding, and multi-object interactions represents a key direction for future work.

\input{sec/X_qwen_questions}

%% file: figures/ff_different_styles.tex
\begin{figure}[h]
    \centering
    \includegraphics[trim={0 92cm 3cm 0}, clip, width=1.0\linewidth]{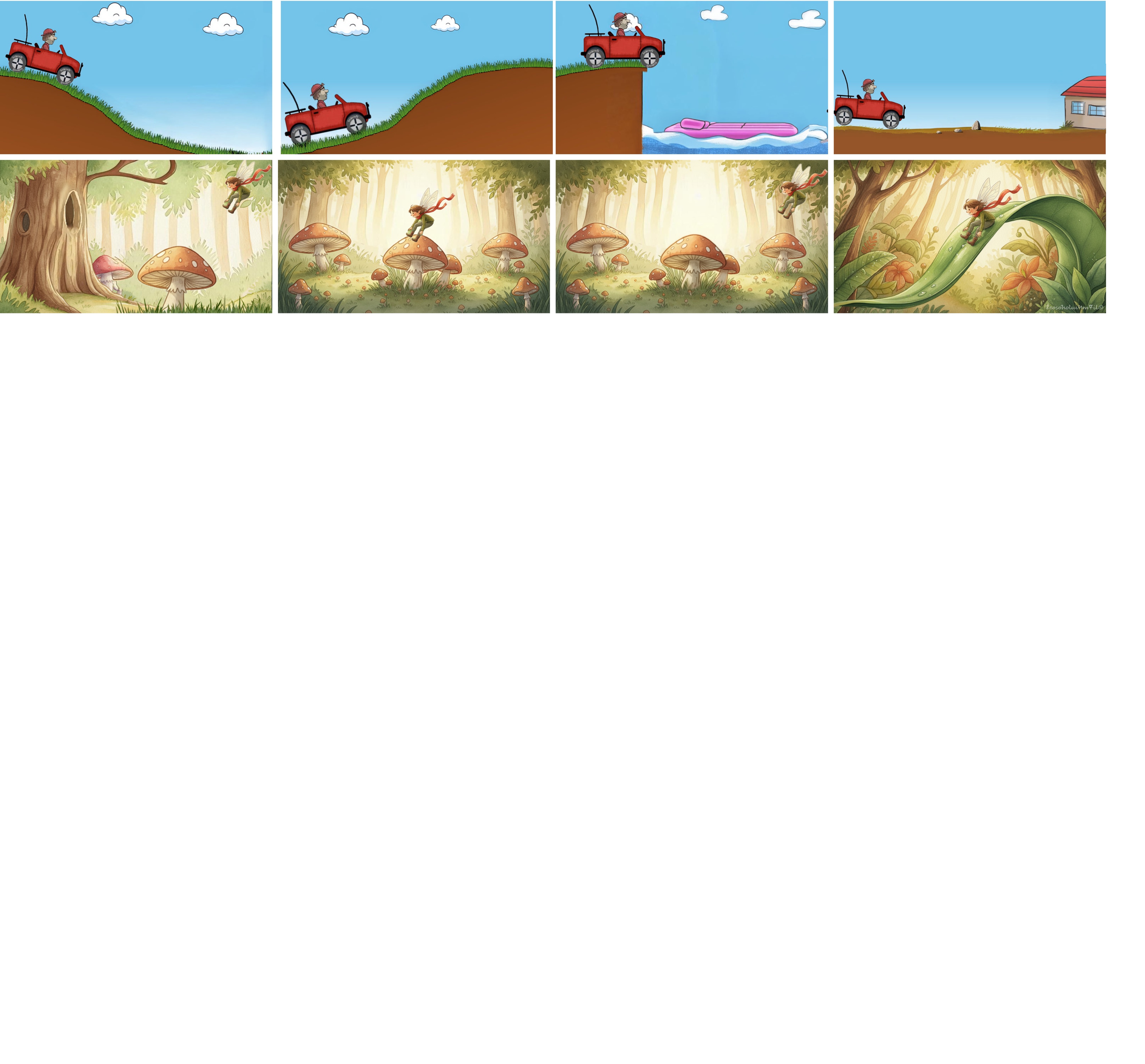}
    \caption{Shows the first frames that were used in the generated video storyline for two different examples}
    \label{fig:storyline_ff}
\end{figure}

%% file: tables/fvmd_scores.tex

\begin{table}[h]
    \centering
    \scriptsize
    \begin{tabular}{l|ccccc}
        \toprule
        Methods & S.M. & F.D. & Opt. & Mag. & Therm. \\
        \midrule
        Base Model (zero-shot) 
        & 4676.9 
        & 3277.9 
        & 3200.0 
        & 1586.8 
        & \cellcolor{best}\textbf{1618.9} \\
        
        Text-only (finetuned) 
        & 3565.8 
        & 1782.6 
        & 4486.4 
        & 1164.0 
        & \cellcolor{second}1736.6 \\
        
        ControlNet(-VLM) 
        & \cellcolor{second}2340.0 
        & \cellcolor{second}1223.9 
        & \cellcolor{best}\textbf{2991.0} 
        & \cellcolor{second}646.2 
        & 2164.0 \\
        
        ControlNet(+VLM) 
        & \cellcolor{best}\textbf{2337.7} 
        & \cellcolor{best}\textbf{1223.1} 
        & \cellcolor{second}3032.9 
        & \cellcolor{best}\textbf{643.7} 
        & 2132.6 \\
        \bottomrule    
    \end{tabular}
    \caption{
    FVMD-based comparison \cite{liu2024frechetvideomotiondistance} of the proposed methods against base model on Physics-IQ benchmark videos.
    }
    \vspace{-0.25cm}
    \label{tab:fvmd_scores}
\end{table}

%% file: figures/physics_iq_qual.tex
\begin{figure}
    \centering
    \includegraphics[trim={0 40cm 33cm 0}, clip, width=1.0\linewidth]{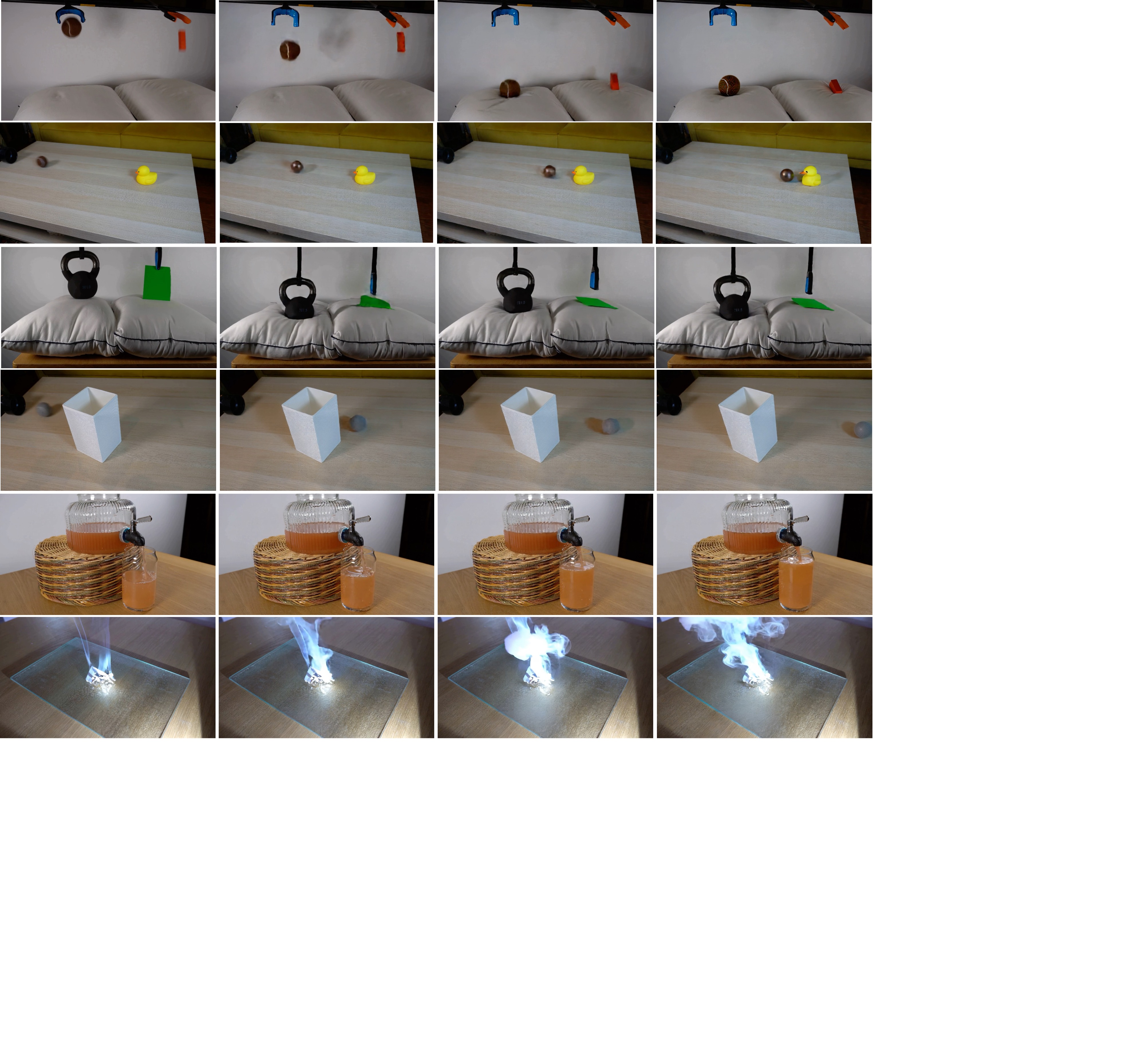}
    \caption{Qualitative results from Physics-IQ~\cite{physicsiq} benchmark.}
    \label{fig:physics_iq_qual}
\end{figure}

%% file: figures/wan22_rebuttal.tex


\begin{figure*}
    \centering
    \includegraphics[trim={0 101cm 7cm 0}, clip, width=1.0\linewidth]{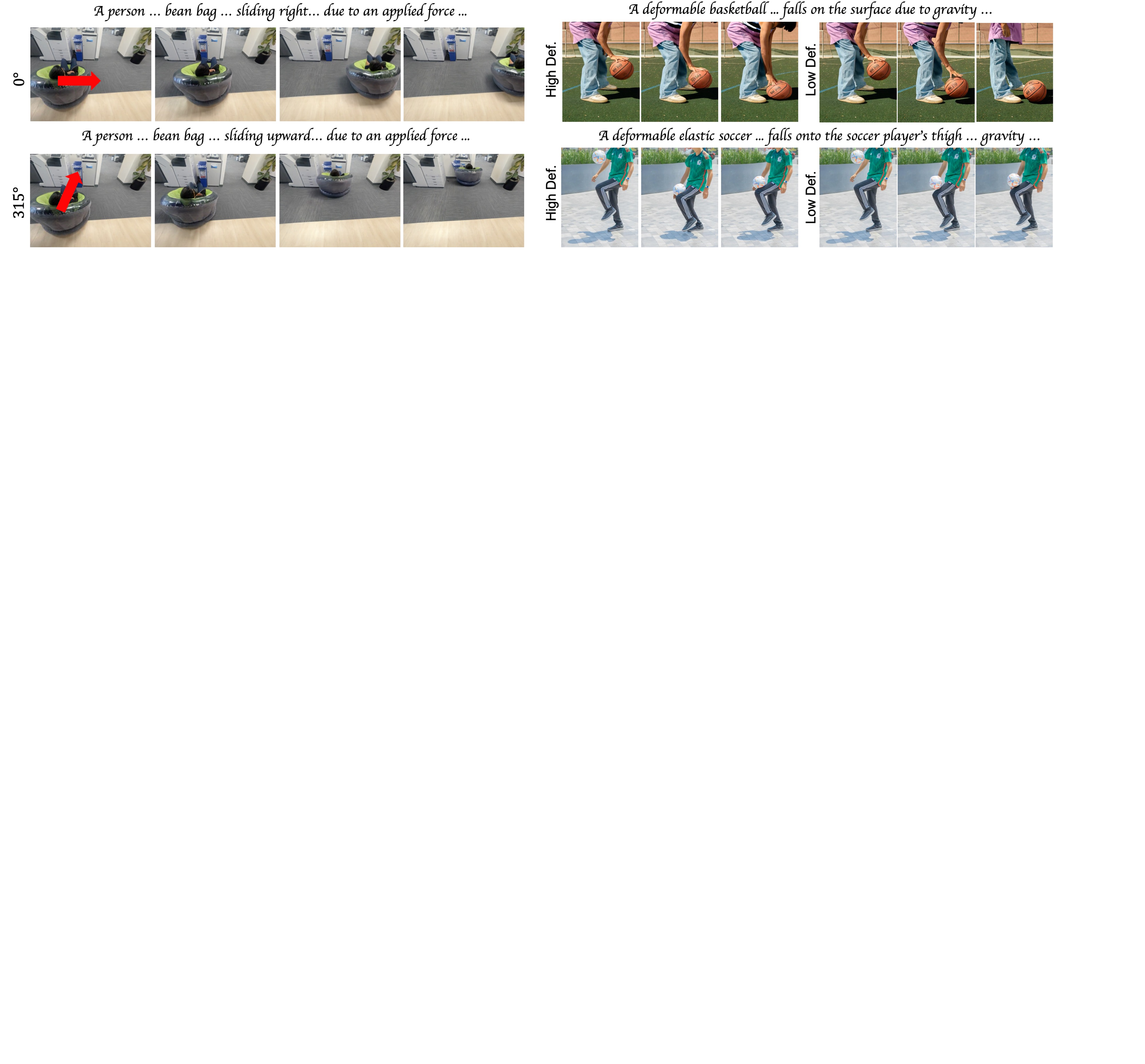}
    \caption{Controllability results from Wan2.2 LoRA model trained with text-only conditioning using the proposed PhyCo dataset.}
    \label{fig:wan22_rebuttal}
    \vspace{-0.3cm}
\end{figure*}

%% file: tables/wan22_physics_iq.tex
\begin{table}[h]
    \centering
    \scriptsize
    \begin{tabular}{l|cccccc}
        \toprule
        Methods & S.M. & F.D. & Opt. & Mag. & Therm. & Avg. \\
        \midrule
        Wan2.2 (zero-shot) & 34.3 & 35.2 & 18.1 & 10.7 & \cellcolor{best}\textbf{36.0} & 30.5 \\
        PhyCo finetuned &  
        \cellcolor{best} \textbf{42.1} & 
        \cellcolor{best}\textbf{37.6} & 
        \cellcolor{best}\textbf{21.9} & 
        \cellcolor{best}\textbf{12.2} & 22.1 &
        \cellcolor{best}\textbf{35.1} \\
        \bottomrule    
    \end{tabular}
    \caption{Quantitative evaluation (120f @ 24FPS) on Physics-IQ benchmark with text only LoRA finetuning on Wan2.2 base model.}
    \vspace{-0.3cm}
    \label{tab:wan22_physics_iq}
\end{table}

%% file: figures/fps_ablation_qual.tex

\begin{figure}[!t]
    \centering
    \includegraphics[trim={0 84cm 28cm 0}, clip, width=1.0\linewidth]{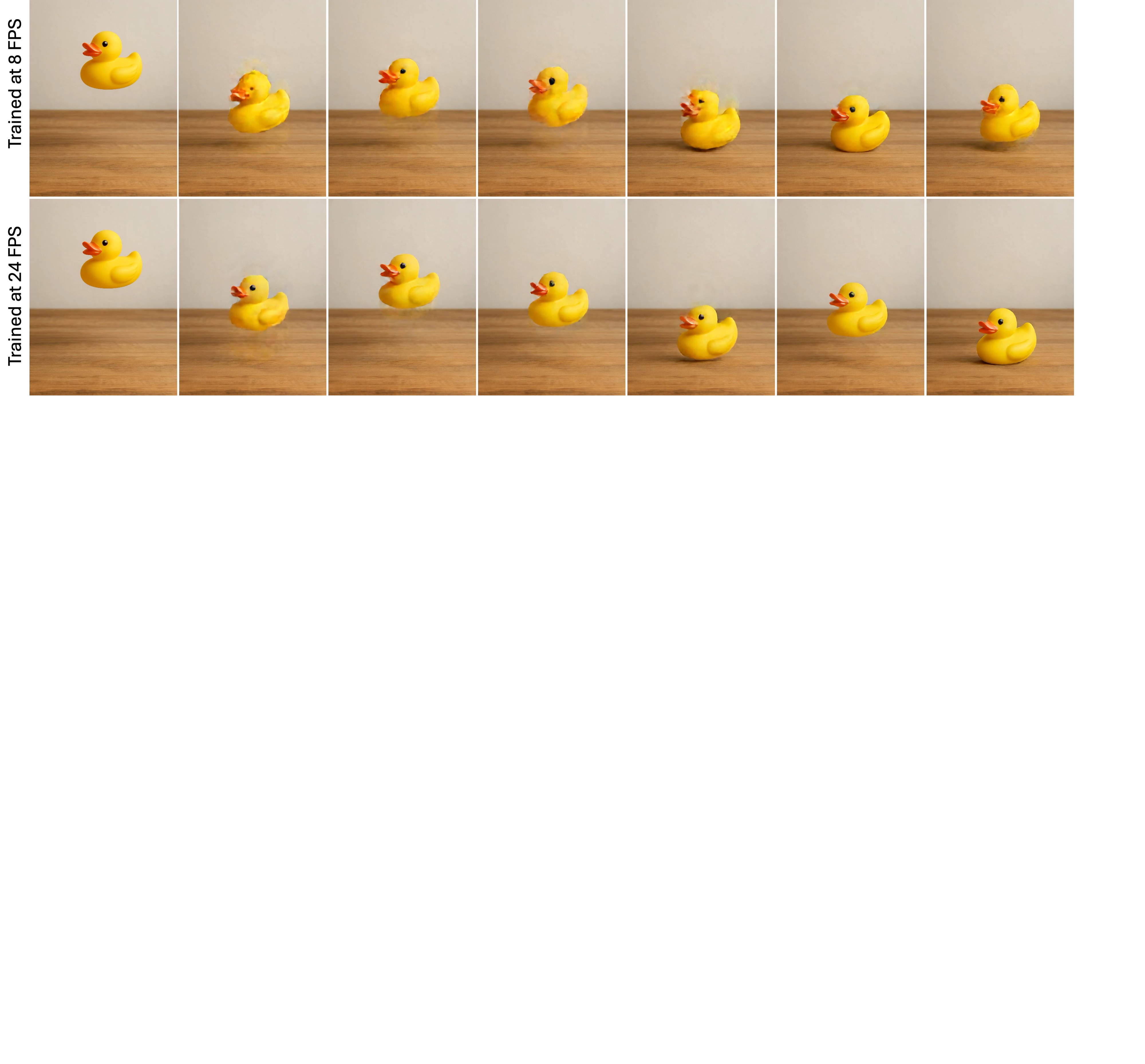}
    \caption{Illustration highlighting the impact of training FPS towards flickering artifacts on Cosmos-Predict base model.}
    \label{fig:fps_ablation_qual}
\end{figure}

%% file: figures/user_study_fig.tex
\begin{figure}
    \centering
    \includegraphics[trim={0 6cm 0cm 0}, clip, width=1.0\linewidth]{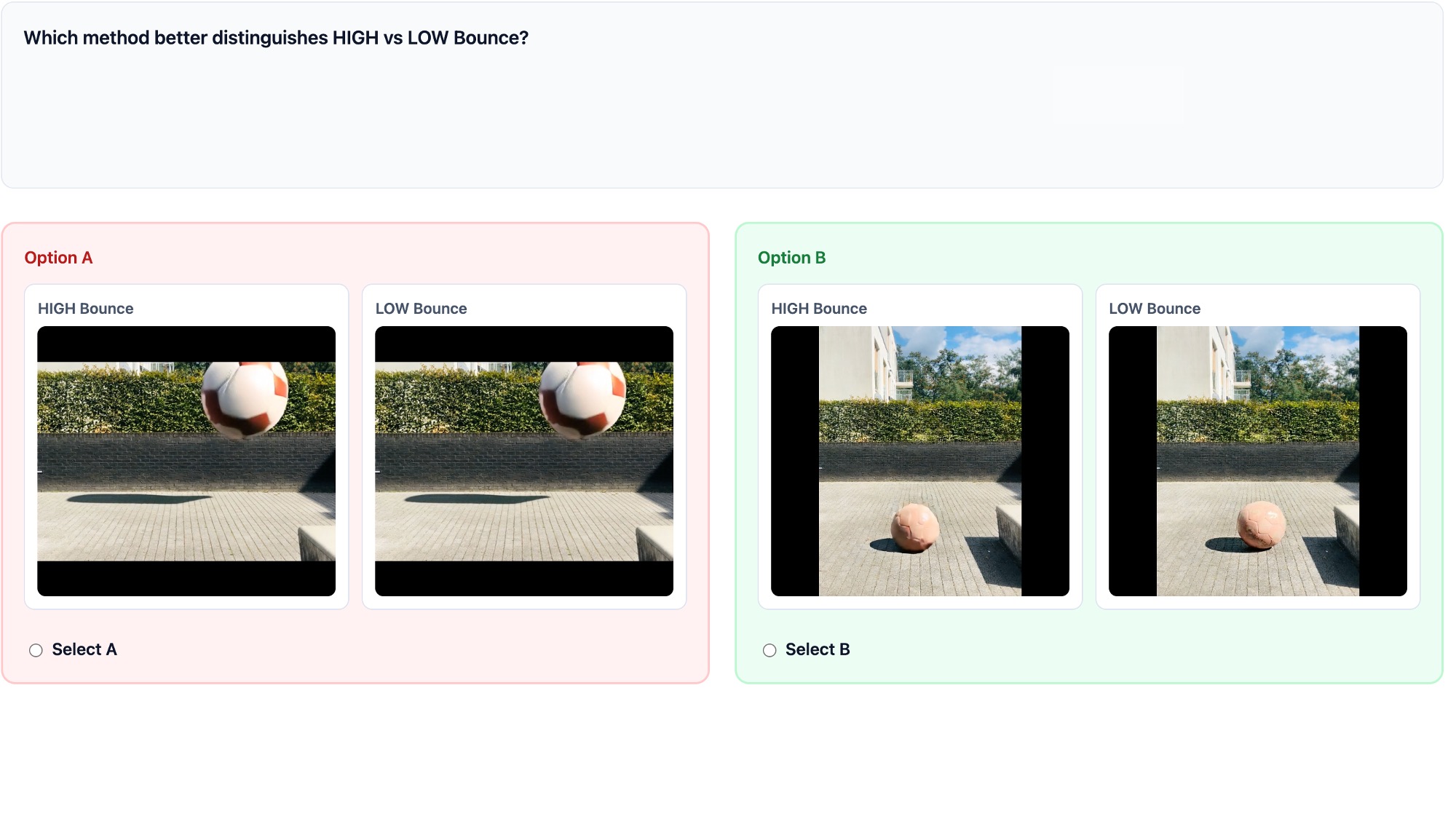}
    \caption{User study interface. Participants were shown two videos for each question and asked to choose which method better expresses the specified physical property.}
    \label{fig:user_study}
\end{figure}

%% file: figures/force_dir_overlay.tex
\begin{figure}
    \centering
    \includegraphics[trim={0 25cm 0cm 0}, clip, width=1.0\linewidth]{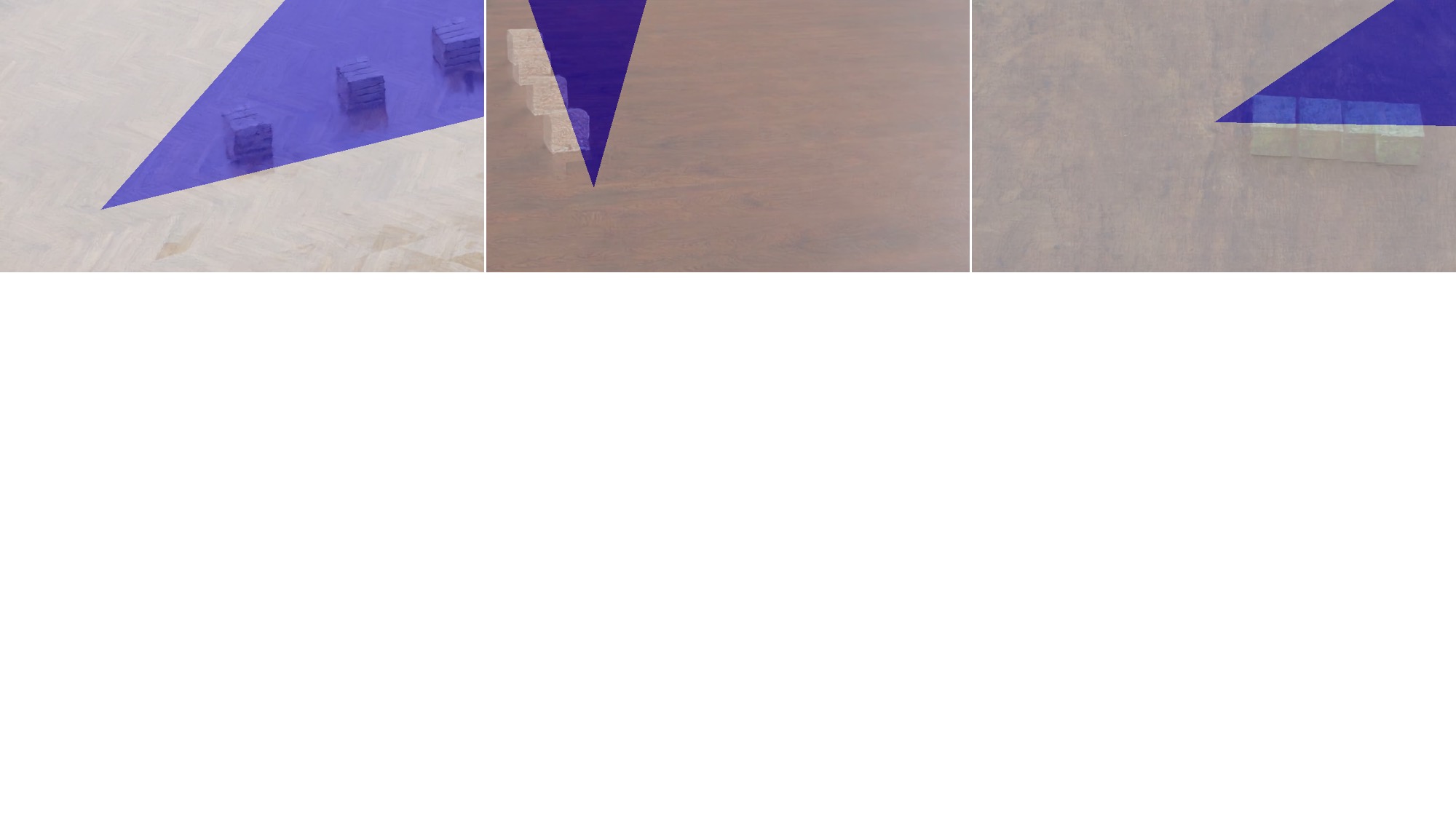}
    \caption{Figure illustrating denoised network outputs with a blue sector overlay indicating the direction of the applied force. The visualization shows the temporal average across video frames.}
    \label{fig:rollout_overlay}
\end{figure}

%% file: sec/X_qwen_questions.tex
\clearpage
\onecolumn

\begin{vlmprompts}[vlm_prompts_box]{VLM Fine-Tuning Prompts}
\footnotesize

\textbf{Force (Sector Adherence)}\\[0.25em]
\textbf{System prompt:}
You are a physics expert analyzing object's motion in a video.\\[0.4em]
\textbf{User prompt:}
I'm showing you a video with a blue highlighted sector region overlaid on all frames.
The blue region is a sector that is a few degrees wide. I want you to carefully
observe the video and answer the following question: Does the object's movement lie
within the blue highlighted sector region? Your answer should be 'Yes' or 'No' only.

\bigskip\hrule\bigskip

\textbf{Force Range Yes/No}\\[0.25em]
\textbf{System prompt:}
You are a physics expert analyzing object's motion in a video.\\[0.4em]
\textbf{User prompt:}
I'm showing you a video with an object sliding on a surface. I want you to carefully
observe the object's motion in thevideo and answer the following question: Is the
force applied to the object between \{min\_value\} and \{max\_value\}? Your answer
should be 'Yes' or 'No' only.

\bigskip\hrule\bigskip

\textbf{Friction Range Yes/No}\\[0.25em]
\textbf{System prompt:}
You are a physics expert analyzing object's motion in a video.\\[0.4em]
\textbf{User prompt:}
I'm showing you a video with an object sliding on a surface. The surface has some
roughness and is only observable based on the object's sliding motion. I want you to
carefully observe the object's motion in thevideo and answer the following question:
Is the friction between the object and the surface between \{min\_value\} and
\{max\_value\}? Your answer should be 'Yes' or 'No' only.

\bigskip\hrule\bigskip

\textbf{Restitution Range Yes/No}\\[0.25em]
\textbf{System prompt:}
You are a physics expert analyzing object's motion in a video.\\[0.4em]
\textbf{User prompt:}
I'm showing you a video with an object bouncing on a surface. The object's bounciness
is observable based on the object's bouncing motion. I want you to carefully observe
the object's motion in the video and answer the following question: Is the
bounciness of the object between \{min\_value\} and \{max\_value\}? Your answer
should be 'Yes' or 'No' only.

\bigskip\hrule\bigskip

\textbf{Deformability Range Yes/No}\\[0.25em]
\textbf{System prompt:}
You are a physics expert analyzing object's motion in a video.\\[0.4em]
\textbf{User prompt:}
I'm showing you a video with a deformable object dropping on a surface. I want you
to carefully observe the object's deformation in the video and answer the following
question: Is the deformability of the object between \{min\_value\} and
\{max\_value\}? Your answer should be 'Yes' or 'No' only.

\bigskip\hrule\bigskip

\textbf{Force Magnitude JSON}\\[0.25em]
\textbf{System prompt:}
You are a physics expert analyzing object's motion in a video. You must respond in
valid JSON format only.\\[0.4em]
\textbf{User prompt:}
I'm showing you a video with an object sliding on the surface. Carefully observe the
object's motion and estimate the magnitude of the force applied to the object.
Respond with a JSON object in this exact format: \{"value": X\} where X is a number
between 0 and 1.

\bigskip\hrule\bigskip

\textbf{Friction Magnitude}\\[0.25em]
\textbf{System prompt:}
You are a physics expert analyzing object's motion in a video. You must respond in
valid JSON format only.\\[0.4em]
\textbf{User prompt:}
I'm showing you a video with an object sliding on a surface. The surface has some
roughness observable from the object's sliding motion. Ignore the object's visual
appearance and focus on its motion to estimate the friction coefficient between the
object and the surface. Respond with a JSON object in this exact format:
\{"value": X\} where X is a number between 0 and 1.

\bigskip\hrule\bigskip

\textbf{Restiution Magnitude}\\[0.25em]
\textbf{System prompt:}
You are a physics expert analyzing object's motion in a video. You must respond in
valid JSON format only.\\[0.4em]
\textbf{User prompt:}
I'm showing you a video with an object bouncing on a surface. The object's
bounciness is observable from its bouncing motion. Ignore the object's visual
appearance and focus on its motion to estimate the coefficient of restitution
between the object and the surface. Respond with a JSON object in this exact format:
\{"value": X\} where X is a number between 0 and 1.

\bigskip\hrule\bigskip

\textbf{Deformation Magnitude}\\[0.25em]
\textbf{System prompt:}
You are a physics expert analyzing object's motion in a video. You must respond in
valid JSON format only.\\[0.4em]
\textbf{User prompt:}
I'm showing you a video with a deformable object dropping on a surface. Ignore the
object's visual appearance and carefully observe its deformation behavior to
estimate how deformable the object is. Respond with a JSON object in this exact
format: \{"value": X\} where X is a number between 0 and 1.

\label{prompt_box}
\end{vlmprompts}

\twocolumn